\documentclass[journal]{IEEEtran}

\usepackage[utf8]{inputenc}
\usepackage[T1]{fontenc}
\usepackage{cite}

\usepackage{booktabs}
\usepackage{multirow}

\usepackage{authblk}
\usepackage{graphicx}
\usepackage{makecell}
\usepackage{subcaption}
\usepackage{amsmath} 
\usepackage{amssymb}  

\usepackage{amsthm}                      
\newtheorem*{remark}{Remark}

\begin{document}

\title{\fontsize{20}{21}\selectfont
Kinematic Reward, Measured Effort: Hardware-Validated RL Morphology Optimization of a Planar 2R Manipulator
}

\author{Gunesh Gupta$^{1}$ Arvind Kumar Mishra$^{2}$ and Sohom Chakrabarty$^{2}$%
\thanks{$^{1}$Gunesh Gupta is an independent researcher, Haridwar, India {\tt\small guneshguptag@gmail.com}}%
\thanks{$^{2}$Arvind Kumar Mishra and Sohom Chakrabarty are from the Department of Electrical Engineering, Indian Institute of Technology Roorkee, Roorkee 247667, India  {\tt\small arvind\_m@ee.iitr.ac.in, sohom@ee.iitr.ac.in}}%
}

\hyphenation{op-tical net-works semi-conduc-tor}

\maketitle
\begin{abstract}

Yoshikawa's Manipulability Index is generally used as a classical measure to define 
an optimal morphology of a manipulator. However, any such optimization is 
almost always studied as a purely kinematic quantity in simulation and is
rarely validated on hardware. We present a 
reinforcement-learning framework with RSL-RL PPO that optimizes the morphology of a 2R SCARA-type
manipulator arm with Yoshikawa’s kinematic manipulability index as a core reward,
and the analytically derived closed-form optimum for the
circular task is used to verify that our method recovers it. Then the same method is applied to custom elliptical and rectangular paths having no analytical optimum. 
The RL-optimized morphologies are then transferred to a 2R hardware manipulator to validate its effect in reduction of joint load in the hardware.
Compared to the worst morphology, the optimum morphology delivered by our method reduces second joint load by $38.4 \pm 0.6\%$ for the ellipse path and $41.3 \pm 0.3\%$ for the rectangle, and
first joint load by $15\%$.

\end{abstract}

\begin{IEEEkeywords}
Manipulators, Reinforcement learning, Manipulability, Robot kinematics, Morphology Optimization, RL-based design.
\end{IEEEkeywords}

\IEEEpeerreviewmaketitle

\section{Introduction}

An effective morphology for a manipulator is inherently
task-dependent. The link lengths and their proportions determine how
dexterously the arm moves and how much effort a given task requires.
Yoshikawa's index $w=\sqrt{\det(JJ^{\mathsf T})}$, the volume of the
manipulability ellipsoid~\cite{yoshikawa1985}, quantifies this.
Maximizing its closed form $w=L_1 L_2\,|\sin\theta_2|$ for a base-centered
circle yields equal link lengths $L_1=L_2=R/\sqrt{2}$ and an orthogonal
configuration $\theta_2=90^\circ$ for a 2R planar manipulator. Most practical tasks, such as off-center
or non-circular paths, have no closed-form optimum even for the 2R planar case and must be solved
numerically by classical global optimizers such as particle-swarm,
Bayesian, or evolutionary methods~\cite{kennedy1995,jones1998,hansen2006}.
More recently, reinforcement learning~\cite{sutton2018} has  also been
applied to this morphology-selection problem, posed analytically
as a single-shot selection in simulation~\cite{mishra2025morphology}.
Classical methods have optimized task-specific link lengths, for instance to minimize energy over a prescribed SCARA path~\cite{mashali2020scara}, and task-driven co-design jointly optimizes morphology and control~\cite{schneider2024codesign}. The motivation is economical: a manipulator dedicated to a repetitive planar operation---adhesive dispensing, planar inspection, sorting along fixed routes, etc---is most efficient when its links are sized to that task at design time rather than left general-purpose. This motivation is aligned with the United Nations SDG~8 to create resource-efficient automation.

In~\cite{mishra2025morphology}, morphology selection has remained entirely
kinematic and simulation-based. A morphology that maximizes $w$ is chosen as
optimal, but it remained unverified how this morphology impacts the physical arm by noting its effect on any dynamic quantity. 
In this paper, we study the effect of kinematically optimized morphology on the reduction of joint efforts in the manipulator. For constant-speed
tracing the joint torque sustaining the commanded motion scales inversely
with manipulability (see Section~\ref{sec:bridge}), so a purely kinematic reward
should favour a lower-effort morphology, a prediction testable on hardware.

The paper closes the sim-to-real gap. We train an RL policy in NVIDIA's Isaac 
Lab from a kinematic reward and validate the resulting morphology on a physical
planar~2R SCARA-style arm built from FeeTech~ST3215 serial-bus servos. Each
servo's on-board load telemetry enables a relative effort measure.

The contributions of this work are as follows:

\begin{enumerate}

 \item \emph{A continuous-control task-specific formulation of the morphology search.} 
  The link lengths are actuated under closed-loop motion and optimized
  with PPO under full rigid-body physics in NVIDIA's Isaac Lab. A length
  budget keeps the search from collapsing to the trivially fully-extended
  arm, so the recovered morphology is a task-specific proportion rather than
  a uniform scaling. RL is the search vehicle, re-run per task like any
  optimizer; at this two-dimensional scale its morphology prediction is cross-checked
  against an exhaustive sweep, while the same formulation could scale the search
  to higher-DOF morphologies where exhaustive enumeration is infeasible.
  \item \emph{Hardware validation of a kinematic morphology criterion
  against measured effort.} On a physical 2R arm built from ST3215
  serial-bus servos, a morphology selected by a purely kinematic
  manipulability objective is shown to measurably lower second-joint 
  motor load on free-space elliptical and rectangular paths, a reduction that holds across ten repeated trials per comparison. Prior kinematic morphology
  studies lack this sim-to-real check.
     
\end{enumerate}

In Section II, the morphology selection problem within the RL framework is described. Section III highlights the inverse relationship between manipulability and joint effort, establishing the kinematic-to-dynamic bridge. Section IV describes the reward function design for the selected tasks. The experiments and results are given in Section V, and the conclusions are drawn in Section VI.

\section{Problem Formulation and RL Framework}
\label{sec:formulation}
\subsection{Planar 2R Kinematics and Manipulability}

The planar two-revolute (2R) manipulator has joint angles
$\theta=[\theta_1,\theta_2]^{\mathsf T}$ and link lengths $L_1,L_2$. Its
end-effector position follows from the forward kinematics
\begin{align}
x &= L_1\cos\theta_1 + L_2\cos(\theta_1+\theta_2),\\
y &= L_1\sin\theta_1 + L_2\sin(\theta_1+\theta_2),
\end{align}
and the corresponding velocity Jacobian is
\begin{equation}
J(\theta)=
\begin{bmatrix}
-L_1 s_1 - L_2 s_{12} & -L_2 s_{12}\\
\;\;L_1 c_1 + L_2 c_{12} & \;\;L_2 c_{12}
\end{bmatrix},
\end{equation}
with $s_1=\sin\theta_1$, $s_{12}=\sin(\theta_1+\theta_2)$, and
$c_1,c_{12}$ defined analogously for cosine. Yoshikawa's manipulability
measure ~\cite{yoshikawa1985}, the volume of the velocity-manipulability
ellipsoid $w=\sqrt{\det\!\big(JJ^{\mathsf T}\big)}$, reduces for the
square non-redundant 2R to
\begin{equation}
w(\theta_2)=L_1 L_2\,\lvert\sin\theta_2\rvert .
\label{eq:w}
\end{equation}

The measure depends on link lengths through the product $L_1 L_2$ and on
configuration through $\lvert\sin\theta_2\rvert$, vanishing at the
extended ($\theta_2=0^\circ$) and folded ($\theta_2=\pm180^\circ$)
singularities. Equation~\eqref{eq:w} is the kinematic quantity the
policy is rewarded for maximizing along the task path; its connection to
joint effort is established in Section~\ref{sec:bridge}.

\subsection{Reference Paths and Workspace}
\label{sec:paths}

\begin{figure}[h]
  \centering
  \includegraphics[width=1.0\columnwidth]{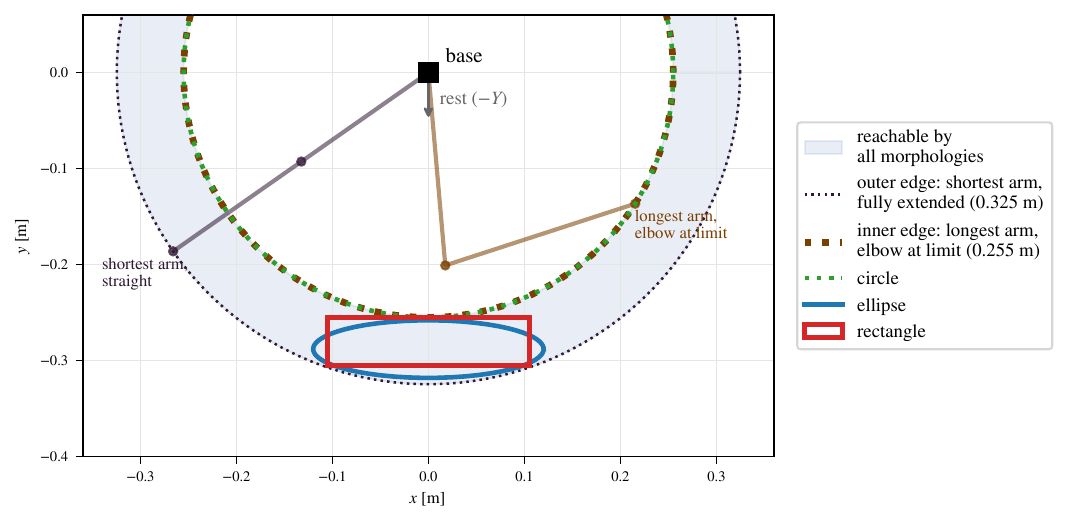}
  \caption{The three task paths: circle (held at radius $R$),
  ellipse, and rectangle. }
  \label{fig:workspace}
\end{figure}

Three task paths are studied, all in the horizontal plane in front of the
base, which rests along negative y-axis at $\theta_1=0$ (Fig.~\ref{fig:workspace}). The
\emph{circle} is centred at the base with radius $R=0.255$~m.
The \emph{ellipse} and \emph{rectangle} are centred at 
$(0,c_y)$ on the rest axis and traced once per episode. 
The ellipse follows
\begin{equation}
p_{\mathrm{ref}}(\phi)=\big(a\cos\phi,\; c_y + b\sin\phi\big),\quad
\phi\in[0,2\pi),
\label{eq:ellipse}
\end{equation}
with $c_y=-0.288$~m, $a=0.12$~m, $b=0.03$~m. The rectangle is the perimeter of
a $2h_L\times 2h_B$ box centred at $(0,c_y)$ ($c_y=-0.28$~m, $h_L=0.105$~m,
$h_B=0.025$~m), traversed counter-clockwise from the right-edge midpoint at a
constant arc-length rate $s(\phi)=(\phi/2\pi)\,P$ with perimeter
$P=2(2h_L+2h_B)=0.52$~m; its corners are exact $90^\circ$ turns with no fillet.
In every case $\phi$ advances uniformly, so the commanded end-effector speed is
approximately constant and all paths lie within the reach of
the admissible arm, whose total length can vary from $0.325$~m
to $0.410$~m (Fig.~\ref{fig:workspace}).

\subsection{Morphology as a Continuously-Actuated Decision Variable}

Morphology optimization is posed as a finite-horizon continuous-control
problem in which the arm tracks a closed reference path while simultaneously
reshaping itself. The link lengths are realized as prismatic degrees of
freedom, so $L_1$ and $L_2$ are actuated quantities rather than fixed
parameters. Each is expressed as a fixed base length plus a bounded slider extension, $L_1\in[0.162,\,0.202]\,\mathrm{m}$ and
$L_2\in[0.163,\,0.208]\,\mathrm{m}$, matching the travel of the physical arm.
The policy commands all four joints through a single action that specifies
scaled position targets for the two revolute joints and the two prismatic
joints,
\begin{equation}
a_t=\big[\,\theta_1^{\mathrm c},\;\theta_2^{\mathrm c},\;L_1^{\mathrm c},\;L_2^{\mathrm c}\,\big]^{\mathsf T},
\end{equation}
where the revolute targets drive the joints that track the moving reference
and the prismatic targets reshape the arm. Revolute and prismatic channels
carry separate scale factors, reflecting their very different physical ranges,
and every commanded morphology lies within the realizable slider travel.
Because morphology and tracking are emitted by the same policy, in simulation the arm
adjusts its length \emph{under motion}. This ensures that the morphology it commits to is the one
that best serves tracking along the entire path rather than at a single
sampled pose.

At each step the policy receives a $19$-dimensional observation. It comprises
the four joint positions and four joint velocities, the two effective link 
lengths, the end-effector tracking error, the path phase encoded as $(\sin\phi,\cos\phi)$
to avoid wrap-around discontinuities, the normalized episode progress $t/T$,
and the previous action. Quantities recoverable by forward kinematics or
fixed by the phase are excluded to keep the observation minimal. Each episode
covers one loop of the reference path, discretized at $60\,\mathrm{Hz}$ over
$5\,\mathrm{s}$ ($T=300$ steps). On reset, the two slider extensions and the path phase $\phi$ are each drawn from an independent uniform distribution over their admissible ranges. So every possible link length pairs $(L_1, L_2)$ is equally likely at initialization. The revolute joints are then initialized by elbow-up inverse kinematics, such that the end-effector begins on the reference trajectory for whatever random morphology was drawn.

\subsection{Objective and Morphology Read-Out}

The policy $\pi_\theta$ is trained with proximal policy
optimization~\cite{schulman2017} to maximize the expected finite-horizon
return
\begin{equation}
J(\pi_\theta)=\mathbb{E}_{\pi_\theta}\!\left[\sum_{t=0}^{T-1} r_t\right],
\label{eq:objective}
\end{equation}
where $r_t$ is the dense per-step reward defined in
Section~\ref{sec:reward}. This reward is purely kinematic and combines a
Gaussian tracking term, the manipulability of~\eqref{eq:w}, and shaping
terms.

Because the link lengths are live actuator outputs, a regularizer reward over 
the final quartile (steps 225-300) of each episode brings the prismatic 
joint movement to rest, resulting in a final morphology output.
Running an inference for this concluded RL policy gives us fixed  
$L_1$ and $L_2$ values for that particular task.
This formulation departs from prior  kinematic morphology studies, 
which select a single fixed morphology and score it by 
averaging manipulability over statically sampled path 
points~\cite{mishra2025morphology}. Here the morphology is exercised under
closed-loop motion and evaluated along the executed trajectory, which is
what makes the downstream comparison of physical joint effort across morphologies
meaningful.

\section{Kinematic-to-Dynamic Bridge}
\label{sec:bridge}
The kinematic reward of Section~\ref{sec:reward} is validated by a dynamic
quantity---measured joint motor load. The two are reconciled by textbook
manipulator dynamics: for constant-speed tracing the joint effort inversly follows
the manipulability $w$ of~\eqref{eq:w}, so the morphology that maximizes $w$
also minimizes tracking effort. This is not a result we set out to
establish but the physics that derives a kinematic reward and predicts the
load pattern seen in Section~\ref{sec:results}.

\subsection{Manipulability Governs Joint Effort}
\label{subsec:bridgeA}
For a non-redundant 2R arm, $\lvert\det J\rvert = w$. Inverting the velocity
relation $\dot x_{\mathrm{ee}} = J(q)\,\dot q$ at a fixed end-effector speed gives
\begin{equation}
\lVert\dot q\rVert \;\propto\; \frac{1}{w},
\label{eq:qdot}
\end{equation}
so a configuration near a singularity ($w$ small) forces large joint rates to
sustain the commanded Cartesian motion. In the horizontal plane the rigid-body
dynamics reduce to $\tau = M(q)\,\ddot q + C(q,\dot q)\,\dot q$, the gravity term
vanishing~\cite{spong2006,siciliano2009}. From the standard two-link
equations~\cite{spong2006}, the dominant configuration-dependent contribution to
the second-joint torque is the centrifugal coupling driven by the first joint,
\begin{equation}
\tau_2 \;\approx\; m_2\,L_1\,L_{c2}\,\sin\theta_2\,\dot\theta_1^{2}
       \;=\; m_2\,\beta\,w\,\dot\theta_1^{2},
\label{eq:bridge}
\end{equation}
where $m_2$ and $L_{c2}$ are the mass and centre-of-mass offset of the second
link, and $\beta=L_{c2}/L_2$ is the dimensionless
centre-of-mass offset of that link, approximately constant across the slider range.
Since, $\dot\theta_1^{2}\propto 1/w^{2}$ from~\eqref{eq:qdot}, it follows that
\begin{equation}
\tau_2 \;\propto\; w^{-p}, \qquad p \approx 1 .
\label{eq:tau-scaling}
\end{equation}

\begin{figure}[h]
  \centering
  \includegraphics[width=\columnwidth]{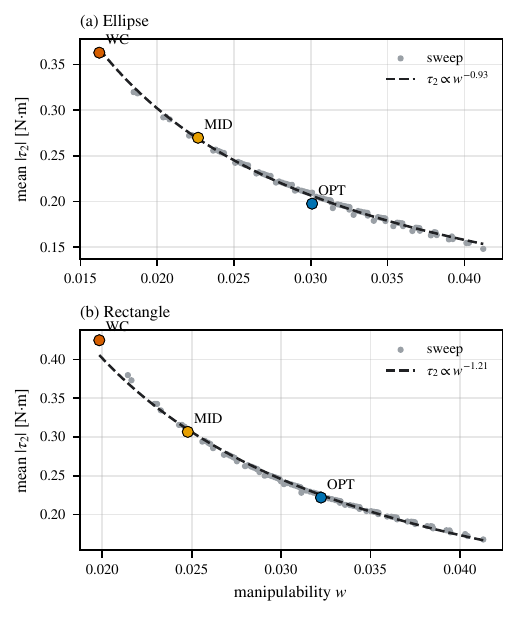}
  \caption{Mean second-joint torque vs.\ manipulability $w$ over a dense
  morphology sweep (simulation, EE mass $+60$~g): ellipse (top), rectangle (bottom),
  with the $\tau_2\propto w^{-p}$ fit and OPT/MID/WC marked
  (from Section~\ref{subsec:sim}).}
  \label{fig:sweep}
\end{figure}

A least-squares fit to the dense morphology sweep (Fig.~\ref{fig:sweep}) gives
$p\approx0.93$ (ellipse) and $p\approx1.21$ (rectangle), close to the predicted
unit exponent, the excess on the rectangle being due to the corner contribution
(see Section~\ref{sec:reward-rect}). The second-joint load is
thus a monotone image of $w$ along the path, which makes the second-joint motor
load the primary hardware discriminator between morphologies. The coupling
vanishes as $m_2\to0$, indicating that on a bare arm $\tau_2$ carries no morphology signal. This motivates a fixed link-2 payload to be mounted for all hardware comparisons
(Section~\ref{subsec:hardware}).

\subsection{Scope of the Bridge}
Equation~\eqref{eq:bridge} assumes free, horizontal-plane motion at approximately
constant end-effector speed. Gravity then acts normal to the plane and does not
load the joints. Within this regime the kinematic measure $w$ predicts
the second-joint load along the path as per \eqref{eq:tau-scaling}.
The load decreases monotonically as the arm lengthens, so the lowest-load
configuration would be the longest admissible arm. The length regularizer
excludes this trivial configuration, leaving the reported optimum as 
the lowest-load morphology the length budget allows within the real-time physics 
simulation environment of NVIDIA Isaac Lab.

\section{Reward Design}
\label{sec:reward}

Each task in NVIDIA Isaac Lab is a finite-horizon tracking episode (Section~\ref{sec:formulation}) in
which the 2R planar manipulator simultaneously tracks a reference path and selects the link lengths $L_1, L_2$ through two prismatic sliders.

\subsection{Analytic Recovery: the Circle}
\label{sec:reward-circle}

The circular task is a closed-form check that RL recovers the classical
Yoshikawa optimum. The end-effector is held on a circle of radius
$R=0.255$~m centred at the base while the agent maximizes a normalized
manipulability term,
\begin{equation}
r_{\mathrm{manip}}
  = \frac{4\,L_1 L_2 \,\lvert \sin\theta_2\rvert}{(L_1 + L_2)^2},
\label{eq:circle-manip}
\end{equation}
which attains its maximum of unity at $L_1 = L_2$, $\theta_2 = 90^\circ$.
A radial Gaussian $\exp\!\big(\!-k\,(r_{ee}-R)^2\big)$, $k = 2000$, holds the
end-effector on the target radius, and a small action-rate penalty
suppresses jerky commands. No length regularizer is needed here. The
constraint $L_1^2 + L_2^2 = R^2$ and the $(L_1+L_2)^2$ normalization together
remove any incentive to grow the links. Since the circle admits a single
static optimal pose rather than a traced path, the episode holds the
end-effector at radius $R$ and reads out the converged morphology directly.
The policy recovers the symmetric configuration predicted in closed form
(Section~\ref{sec:formulation}). The manipulability, radial, and action-rate
terms carry weights $100$, $3.0$, and $0.1$.

\subsection{Trajectory Tasks: the Common Reward Chassis}
\label{sec:reward-chassis}

The ellipse and rectangle have no closed-form optimum. For both, the
end-effector traces the full path, and the reward shares a common chassis of
five terms. The manipulability term reduces to the configuration-only factor
\begin{equation}
r_{\mathrm{manip}} = \lvert \sin\theta_2 \rvert ,
\label{eq:manip-sin}
\end{equation}
dropping the $L_1 L_2$ product so the term rewards elbow configuration alone.
Retaining the product would make longer links monotonically more rewarding
and bias the policy toward full extension. Isolating $\lvert\sin\theta_2\rvert$
instead leaves arm-size control to a separate linear regularizer
$r_{\mathrm{len}} = -(L_1 + L_2)$. This regularizer does more than debias the
reward. Without it the objective is unbounded, and every task collapses to
maximal extension, so the regularizer is what makes the optimum well-posed
and task-dependent. It is also a physical proxy. In the simulation each link body has a fixed mass,
so extending a slider lengthens the arm and moves the existing distal mass
farther from the joint without adding any new material mass. A physically
longer link would weigh more, so $r_{\mathrm{len}}$ stands in for the self-mass
and footprint cost that the fixed-mass simulation model does not capture. Path tracking 
uses a Gaussian on the end-effector error,
$r_{\mathrm{track}} = \exp(-\alpha\lVert p_{ee} - p_{\mathrm{ref}}(\phi_t)\rVert^2)$,
$\alpha = 2500$. Its bandwidth sets the effective error scale. A bandwidth
that is too sharp makes the gradient vanish at the large errors of a
randomized start, while one that is too broad never demands sub-centimetre
tracking.

Two further terms settle the morphology. Because $L_1,L_2$ are actions, the
sliders would otherwise drift rather than commit to a single morphology. A
time-ramped settling penalty
$-\gamma(t)\,(\lvert\dot L_1\rvert + \lvert\dot L_2\rvert)$,
$\gamma(t) = \gamma_{\max}\,(t/T)$, leaves them free early for
inverse-kinematic correction from a randomized pose. It then penalizes slider
motion increasingly, so the morphology is committed by the readout window. A small action-rate penalty $-\beta_a \, \| a_t - a_{t-1} \|^2$, using the
built-in term from Isaac Lab~\cite{mittal2023orbit}, prevents the link lengths from changing drastically between subsequent steps. A constant alive bonus $c_{\mathrm{alive}}$, which
offsets the standing length penalty so the policy does not terminate early to
escape it, and a one-shot terminal penalty $c_{\mathrm{term}}$ on non-timeout
termination complete the reward chassis. The full per-step reward is given in~\eqref{eq:reward-full},
with per-task weights in Table~\ref{tab:reward-traj}. Equation~\eqref{eq:reward-full} completes the reward design for the elliptical trajectory tracking task.
\begin{equation}
\begin{aligned}
r(t) ={}& w_{\mathrm{t}}\,\exp\!\big(\!-\alpha\lVert p_{ee} -
          p_{\mathrm{ref}}(\phi_t)\rVert^2\big)
        + w_{\mathrm{m}}\,\lvert\sin\theta_2\rvert \\
      &- w_{\ell}\,(L_1 + L_2)
        - \gamma(t)\,(\lvert\dot L_1\rvert + \lvert\dot L_2\rvert) \\
      &- w_{\mathrm{a}}\lVert a_t - a_{t-1}\rVert^2
        + c_{\mathrm{alive}}
        + c_{\mathrm{term}}
\end{aligned}
\label{eq:reward-full}
\end{equation}

\subsection{Sharp Corners: Singularity and Chatter Control}
\label{sec:reward-rect}

The rectangular tracking task inherits the full reward chassis of
\eqref{eq:reward-full}, including the length regularizer at the
same weight, and adds two structural terms with two further weights.
Its far corners place the end-effector near maximum reach. There the arm
approaches the outer singularity at $\theta_2 = 0$, and the joint speed
required for constant end-effector speed grows as $1/\lvert\sin\theta_2\rvert$.
A smooth corner-floor term
\begin{equation}
r_{\mathrm{floor}}
  = \frac{\min(\lvert\sin\theta_2\rvert,\,\delta)}{\delta},
  \qquad \delta = 0.5,
\label{eq:corner-floor}
\end{equation}
saturates to unity once $\lvert\sin\theta_2\rvert \ge \delta$ and decays
linearly as the elbow approaches a singular pose. It is applied with a
subordinate weight and acts as a safety floor rather than a competing
objective. It is the path-resolved, worst-point analogue of guarding the
smallest singular value of the Jacobian, which for the planar $2$R is
proportional to $\lvert\sin\theta_2\rvert$ along the path (Yoshikawa, \S4.3).

The sharp corners also destabilize the morphology. The
$1/\lvert\sin\theta_2\rvert$ speed demand at each corner injects a transient
slider command on every pass. Prismatic chatter therefore appears throughout
the loop rather than only late in the episode, and the time-ramped settling
penalty, weak early by construction, cannot suppress it alone. An always-on
slider-speed penalty $-w_s\,(\lvert\dot L_1\rvert + \lvert\dot L_2\rvert)$
suppresses this chatter at every phase. The settling ramp is strengthened
from $\gamma_{\max} = 2.0$ to $6.0$ to give it more authority near the end of
the episode. The action-rate weight is increased from $10^{-3}$ to
$5\times10^{-3}$ to break a residual bang-bang pattern of the sliders against
their travel limits. The smooth ellipse path requires none of these
additions. Its reference has a continuous derivative everywhere, so no corner
forces a $1/\lvert\sin\theta_2\rvert$ speed spike, and the reward chassis of~\eqref{eq:reward-full} is sufficient.

\begin{table}[h]
\centering
\caption{Reward weights for the trajectory tasks, code-verified against
each branch. The ellipse uses the common reward chassis. The rectangle adds two
terms (corner floor, slider speed) and retunes two ($\gamma_{\max}$,
action rate).}
\label{tab:reward-traj}
\begin{tabular}{lcc}
\toprule
Term / role & Ellipse & Rectangle \\
\midrule
manipulability $|\sin\theta_2|$, $w_{\mathrm{manip}}$ & $0.5$ & $0.5$ \\
tracking $w_{\mathrm{t}}$ (bandwidth $\alpha$) & $1.0$ ($2500$) & $1.0$ ($2500$) \\
length reg.\ $w_{\ell}$ & $0.5$ & $0.5$ \\
corner floor $w_{\mathrm{floor}}$ ($\delta$) & --- & $0.2$ ($0.5$) \\
settle ramp $\gamma_{\max}$ & $2.0$ & $6.0$ \\
slider speed $w_s$ & --- & $0.3$ \\
action rate $w_{\mathrm{a}}$ & $10^{-3}$ & $5\times10^{-3}$ \\
alive / terminal & $0.01 / {-}5.0$ & $0.01 / {-}5.0$ \\
\bottomrule
\end{tabular}
\end{table}

\begin{remark}
Every reward term is kinematic, so the effort reduction reported in
Section~\ref{sec:results} follows from the kinematic objective rather 
than from any explicit dynamic shaping. Manipulability alone does not 
decide link lengths, so path geometry and closed-loop tracking settle the choice.
\end{remark}

\section{Experiments and Results}
\label{sec:results}

\subsection{Setup}
\label{subsec:setup}

Policies are trained in NVIDIA Isaac Lab on the PhysX engine using PPO from
the \texttt{rsl\_rl} library~\cite{mittal2023orbit,rudin2022learning}. All
four joints are driven by implicit position actuators with stiffness $150$,
damping $5.0$, an effort limit of $2.5$~N$\cdot$m, and a velocity limit of
$10$~rad/s. The simulation steps are at $60$~Hz. Each policy is trained on a
single NVIDIA GeForce RTX~5060 GPU with $4096$ parallel environments and
converges within roughly $3200$ iterations. A full $5000$-iteration run
completes in $10$--$15$~min of wall-clock time. The high environment count is
what makes the converged morphology seed-independent
(Section~\ref{subsec:hardware}). A one-shot morphology selection would be
seed-insensitive by construction, whereas the temporal formulation here
requires sufficient parallel sampling. Gravity is enabled but acts normal to
the horizontal motion plane, consistent with the free-space, in-plane regime
of Section~\ref{sec:bridge}.

\begin{figure}[h]
    \centering
    \begin{subfigure}{0.32\columnwidth}
        \includegraphics[width=\linewidth]{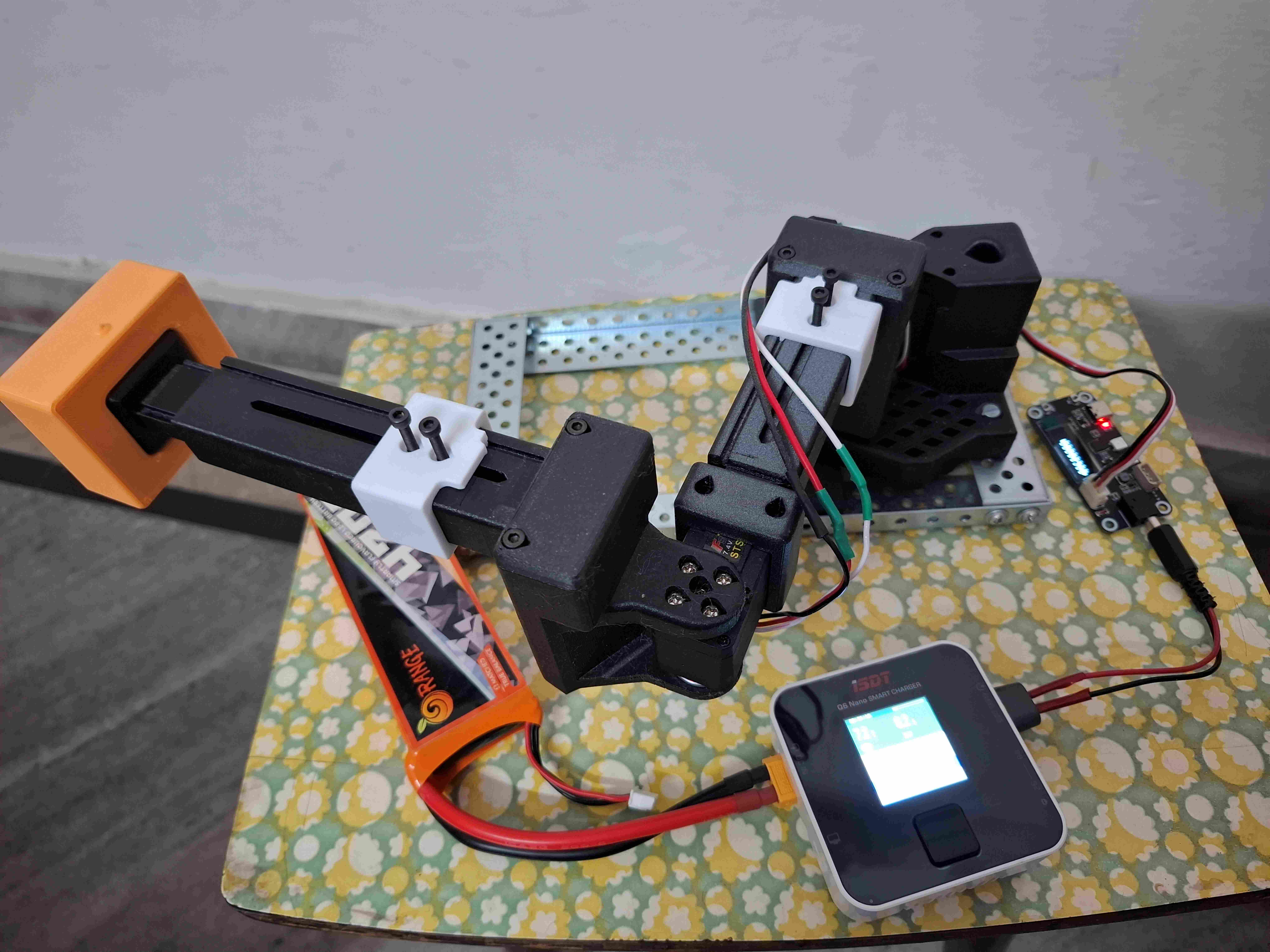}
        \caption{}\label{fig:setup_a}
    \end{subfigure}\hfill
    \begin{subfigure}{0.32\columnwidth}
        \includegraphics[width=\linewidth]{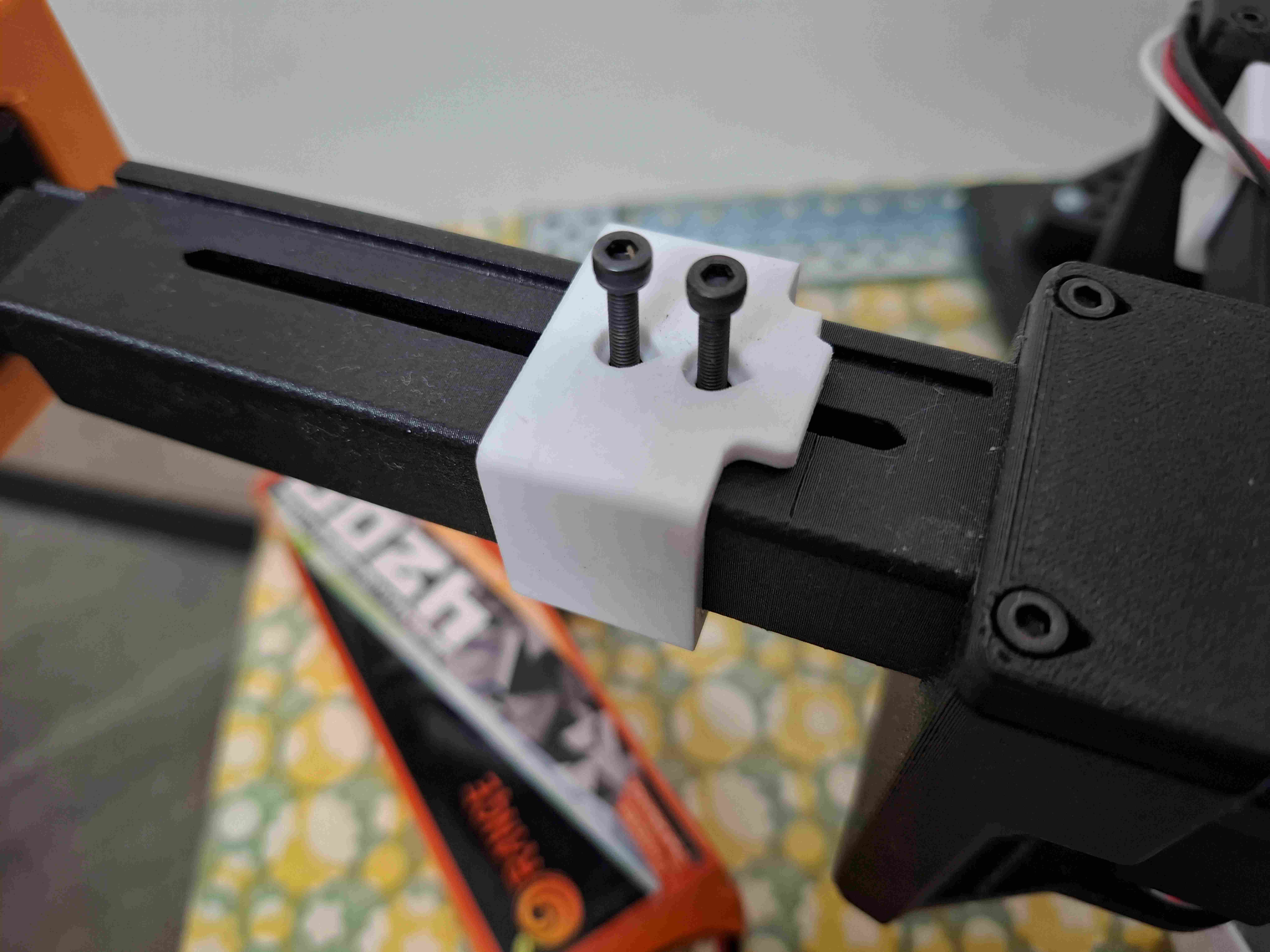}
        \caption{}\label{fig:setup_b}
    \end{subfigure}\hfill
    \begin{subfigure}{0.285\columnwidth}
        \includegraphics[width=\linewidth]{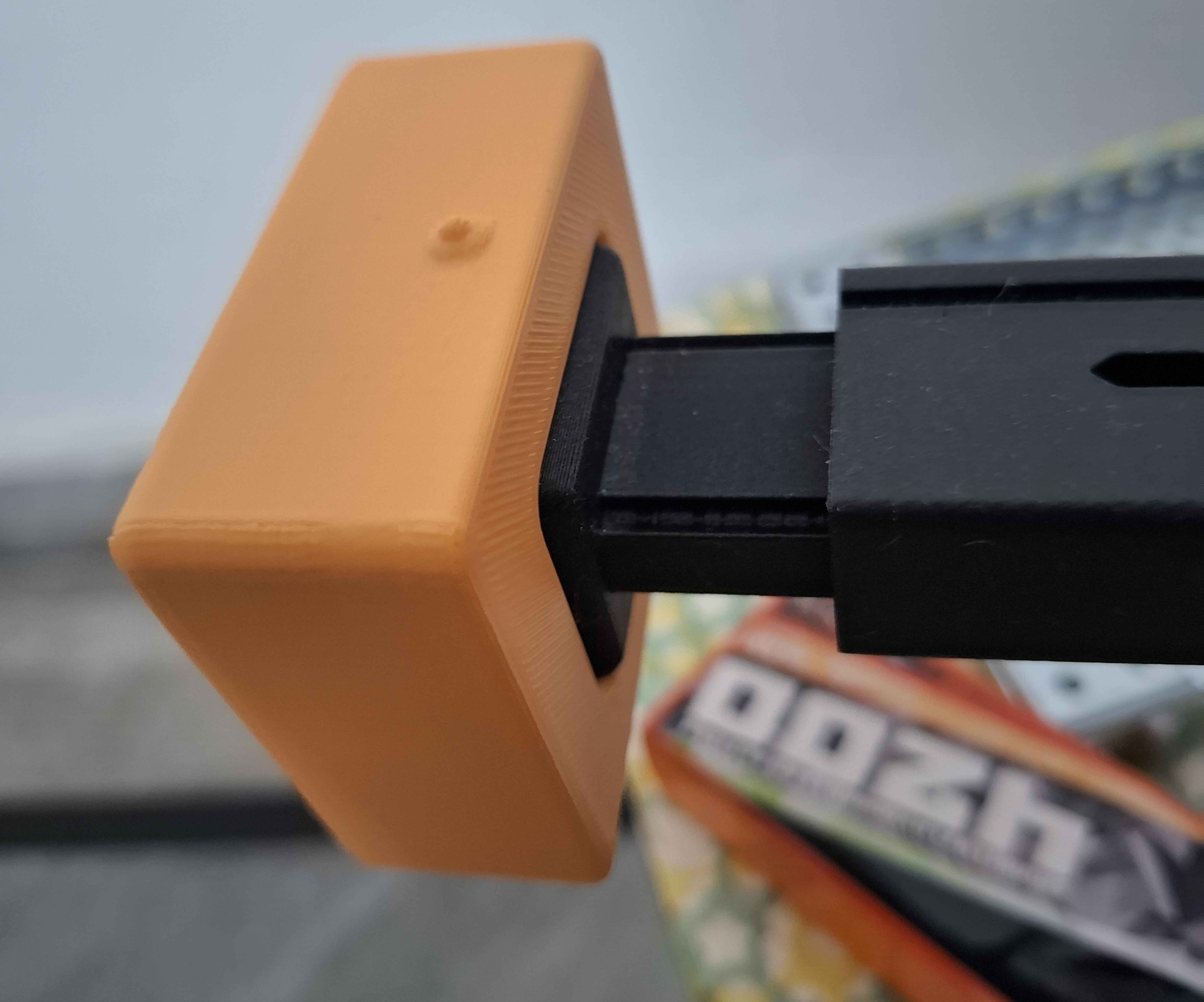}
        \caption{}\label{fig:setup_c}
    \end{subfigure}
    \caption{Hardware platform: (a) Planar 2R arm; (b) prismatic link stage; (c) the $60$~g end-effector payload.}
    \label{fig:full_hardware_setup}
\end{figure}

The physical platform (Fig.~\ref{fig:full_hardware_setup}) is a planar 2R arm
driven by two FeeTech ST3215 serial-bus servos. Each link length is set by a
screw stage and held by a magnetic lock that preserves link inertia. The
morphology of the hardware planar 2R arm is fixed per trial, so the search is done in simulation and the
selected morphology is then built and measured. A $60$~g block is mounted at
the end-effector (EE) for all runs, as indicated in Section ~\ref{subsec:bridgeA}. Joint effort is read from each servo as load
(signed PWM duty) and reported as a relative ranking only, since the torque
constant is uncalibrated.

Each comparison traces the path for two $3$~s loops at $60$~Hz, using an
elbow-up inverse-kinematics command and a trapezoidal phase profile. The arm
is pre-positioned at the start and settled before logging, and statistics use the steady-state window (steps $5$--$350$), and each morphology is measured ten times under identical conditions, with all hardware values reported as the mean over the ten runs and one standard deviation as the uncertainty. The primary metric is the mean
second-joint load $\overline{|j_2|}$, with the secondary metric being the first-joint load $\overline{|j_1|}$. 
Mean tracking error accompanies every comparison
as a fairness gate, so a load reduction counts only when OPT tracks at least
as accurately as the baseline. Three morphologies are compared per task. The
first is the RL-selected optimum OPT (ellipse $L_1{=}0.202$, $L_2{=}0.163$~m;
rectangle $L_1{=}0.162$, $L_2{=}0.208$~m).
The second is a worst-case low-manipulability morphology WC with both sliders
retracted ($L_1{=}0.162$, $L_2{=}0.163$~m). The third is an intermediate point
MID ($L_1{=}0.162$, $L_2{=}0.178$~m) read off the simulated load curve.

\subsection{Simulation}
\label{subsec:sim}

On the analytically solvable circular task the policy recovers the known
optimum. It converges to equal link lengths $L_1{=}L_2{\approx}0.179$~m with
an orthogonal joint configuration ($\theta_2{\approx}90^\circ$), in agreement
with the closed-form $R/\sqrt{2}$ solution. The circle is traced only as an
arc on the physical arm and is therefore reported in simulation only.

For the ellipse and rectangle, the RL-selected OPT
sits high on the manipulability axis ($\bar w=0.030$ ellipse Fig.~\ref{fig:trace_ell_w}, $0.032$
rectangle Fig.~\ref{fig:trace_rect_w}) and WC at the low end ($\bar w=0.016$, $0.018$), with MID between
them. Relative to WC, OPT lowers mean $|\tau_2|$ by $45.5\%$ on the ellipse
and $47.8\%$ on the rectangle. The corner singularity appears in simulation as
a sharp rise in second-joint torque, the rectangle worst case peaking near
$1.9$~N$\cdot$m. On hardware the same corner drives the servo into saturation,
as discussed below.

\vspace{-2pt}
\subsection{Hardware Validation}
\label{subsec:hardware}

\begin{figure}[h]
  \centering
  \includegraphics[width=0.9\columnwidth]{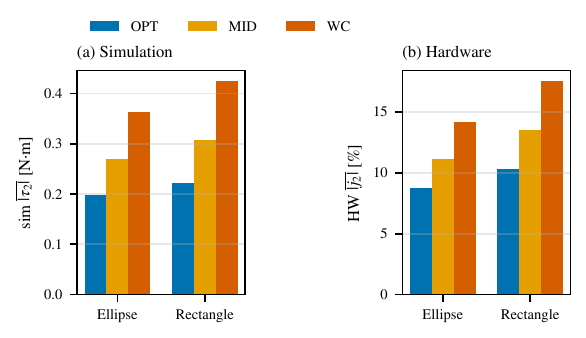}
  \caption{OPT/MID/WC comparison: (a) simulated mean second-joint torque
  [N$\cdot$m]; (b) measured hardware second-joint load [\%, relative, averaged over 10 runs]. Both
  rank OPT$\,<\,$MID$\,<\,$WC (Section~\ref{subsec:hardware}).}
  \label{fig:bars}
\end{figure}

\begin{figure}[h]
  \centering
  \begin{subfigure}[h]{0.9\columnwidth}
    \includegraphics[width=\linewidth]{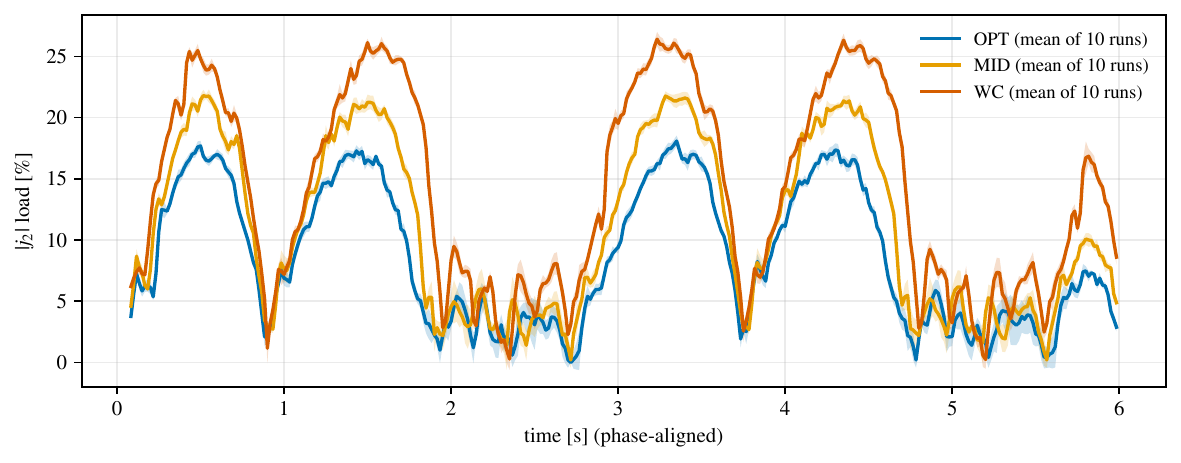}
    \caption{Ellipse: $|\bar{j}_2|$ load [\%]}
    \label{fig:trace_ell_load}
  \end{subfigure}\hfill
  \begin{subfigure}[h]{0.9\columnwidth}
    \includegraphics[width=\linewidth]{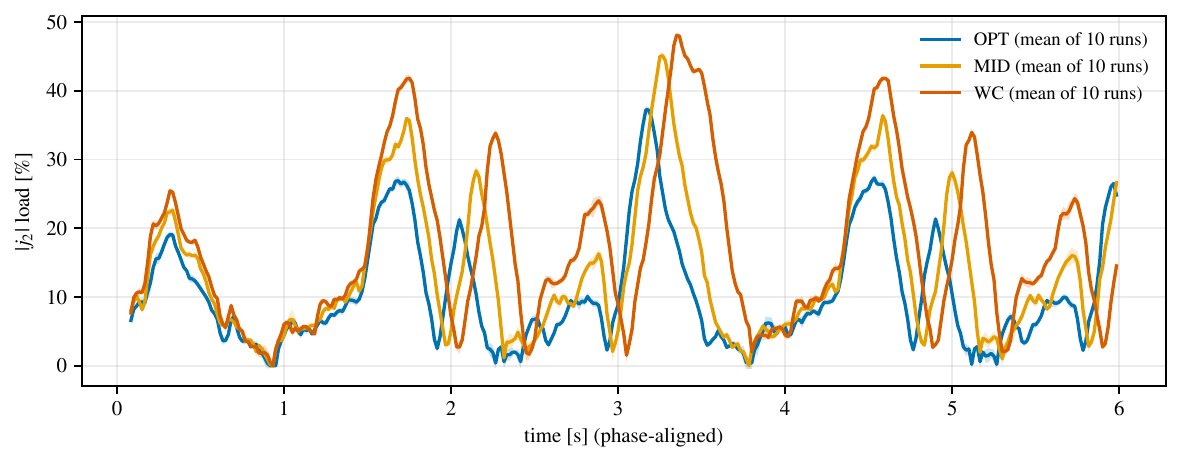}
    \caption{Rectangle: $|\bar{j}_2|$ load [\%]}
    \label{fig:trace_rect_load}
  \end{subfigure}
    \begin{subfigure}[h]{0.9\columnwidth}
    \includegraphics[width=\linewidth]{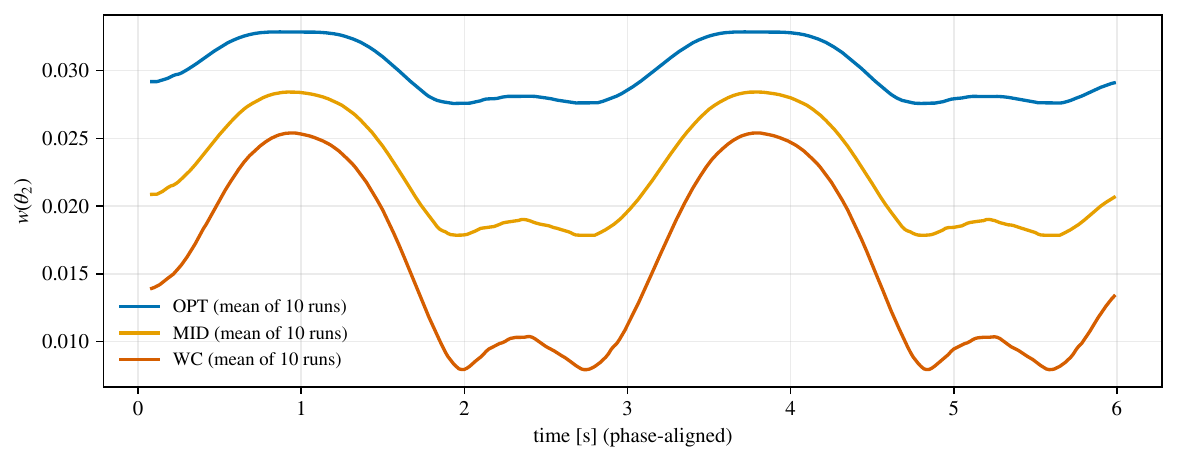}
    \caption{Ellipse: $w(\theta_2)$ along path}
    \label{fig:trace_ell_w}
  \end{subfigure}\hfill
  \begin{subfigure}[h]{0.9\columnwidth}
    \includegraphics[width=\linewidth]{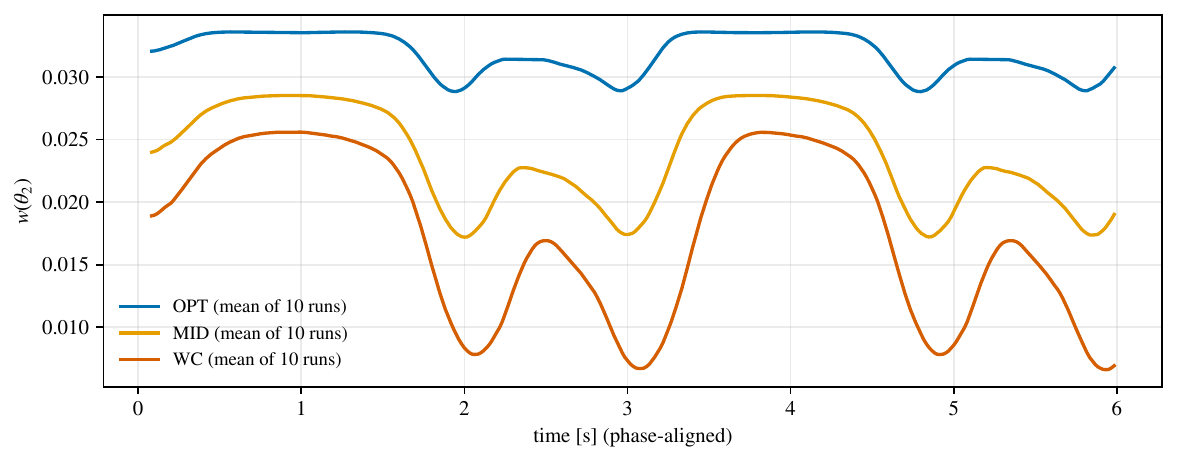}
    \caption{Rectangle: $w(\theta_2)$ along path}
    \label{fig:trace_rect_w}
  \end{subfigure}

  \caption{Plots of $|\bar{j}_2|$ load \% and $w(\theta_2)$ over 2 loops averaged across 10 runs.}
  \label{fig:hw_traces}
\end{figure}

The simulation ranking is reproduced on hardware. For both tasks the
RL-selected morphology lowers the measured second-joint load relative to the
WC baseline throughout the loop (Fig.~\ref{fig:trace_ell_load} and Fig.~\ref{fig:trace_rect_load}). Load $\overline{|j_2|}$
falls by $38.4 \pm 0.6\%$ on the ellipse ($8.74$ vs $14.18$) and by $41.3 \pm 0.3\%$ on the rectangle ($10.29$ vs $17.53$), with MID intermediate, matching the monotonic
ordering of Fig.~\ref{fig:sweep} (Fig.~\ref{fig:bars};
Table~\ref{tab:summary}). 
\begin{table}[h]
  \centering
  \caption{Per-morphology summary. Hardware values are the mean over ten repeated runs (one standard deviation shown), EE mass $+60$~g both simulation and hardware.}
  \label{tab:summary}
  \setlength{\tabcolsep}{4pt} 
  \begin{tabular}{llccccc}
    \toprule
    Task & & $\bar w$ 
    & \makecell{$\min$ \\ $|\sin\theta_2|$} 
    & \makecell{Sim $\overline{|\tau_2|}$ \\ {}[N$\cdot$m]} 
    & \makecell{HW $\overline{|j_2|}$ \\ {}[\%]} 
    & \makecell{$\overline{|\text{track}_{\text{error}}|}$ \\ {}[mm]} \\
    \midrule
    \multirow{3}{*}{Ellipse}
      & OPT & 0.030 & 0.84 & 0.20 & 8.74 $\pm$ 0.08 & 16.6 $\pm$ 0.0 \\
      & MID & 0.022 & 0.62 & 0.27 & 11.10 $\pm$ 0.05 & 17.8 $\pm$ 0.0 \\
      & WC & 0.016 & 0.30 & 0.36 & 14.18 $\pm$ 0.02 & 19.1 $\pm$ 0.0 \\
    \midrule
    \multirow{3}{*}{Rect.}
      & OPT & 0.032 & 0.86 & 0.22 & 10.29 $\pm$ 0.06 & 18.6 $\pm$ 0.0 \\
      & MID & 0.024 & 0.60 & 0.31 & 13.53 $\pm$ 0.04 & 21.6 $\pm$ 0.0 \\
      & WC & 0.018 & 0.25 & 0.42 & 17.53 $\pm$ 0.03 & 30.2 $\pm$ 0.1 \\
    \bottomrule
  \end{tabular}
\end{table}
These reductions are obtained at equal or better
tracking, i.e. $16.6$~mm vs $19.1$~mm tracking error on the ellipse and $18.6$~mm vs $30.2$~mm tracking error Table~\ref{tab:summary} on the
rectangle, so the saving does not come at the cost of accuracy. The larger
tracking gap on the rectangle reflects the retracted arm saturating at the
sharp corners, where it approaches the servo's $1.91$~N$\cdot$m rated torque. The first-joint load also decreases, by 15.7\% on the ellipse and 
15.1\% on the rectangle. The selected morphology also stays well clear of the outer
singularity. The minimum $|\sin\theta_2|$ over the rectangle path 
is $0.86$ for OPT against $0.25$ for the WC, a direct 
consequence of the corner-floor term in the reward.

Three further points support the comparison. First, the corner-spike ratio
(peak-to-mean load) does not discriminate morphology ($2.78$ OPT vs $2.76$ WC). It is
path-driven, so absolute peak load and minimum $|\sin\theta_2|$ are the
appropriate corner metrics. Second, at $4096$ environments every pre-committed 
seed converges to the same morphology, indicating that the selected morphology is seed-independent. Third, across the ten hardware repeats of 
each comparison, the run-to-run variation is small relative to the 
morphology gap (Table~\ref{tab:summary}), so the measured load ranking 
is unambiguous.

\section{Conclusion}

\label{sec:conclusion}
A purely kinematic manipulability reward suffices to select task-appropriate
link proportions for a planar 2R manipulator, and the selected morphology
measurably reduces joint effort on real hardware. Trained with PPO under full
rigid-body physics, the policy recovers the closed-form Yoshikawa optimum on
the circular task and yields distinct, mirror-related morphologies for the
elliptical and rectangular paths. On a physical ST3215-servo arm these
morphologies lower the measured second-joint load by $38.4 \pm 0.6\%$ (ellipse) and $41.3 \pm 0.3\%$ (rectangle) at equal or better tracking. This reproduces the
second-joint-dominant load asymmetry predicted by the torque--manipulability
relation and the scaling $\tau_2\propto w^{-p}$ ($p\approx1$). The
study is confined to free-space, horizontal-plane, constant-speed motion and
reports relative effort rankings rather than absolute torque. Calibrated absolute torque, acceleration-dominated tasks, and higher-DOF arms remain open for exploration in future work.

\bibliographystyle{IEEEtran}
\bibliography{references} 

\end{document}